# Automated Machine Learning in Radiomics: A Comparative Evaluation of Performance, Efficiency and Accessibility

## Author Block

Jose Lozano-Montoya[1,2], Emilio Soria-Olivas[3], Almudena Fuster-Matanzo[2], Angel Alberich-Bayarri[2], Ana Jimenez-Pastor[2]

[1]University of Valencia, Valencia, Spain

[2]Research & Frontiers in AI Department, Quantitative Imaging Biomarkers in Medicine, Quibim SL, Valencia, Spain

[3]Intelligent Data Analysis Laboratory, IDAL, University of Valencia, Valencia, Spain

## Corresponding Author

Jose Lozano-Montoya, joselozano@quibim.com

Address: Europa Building, Aragon Avenue, 30, 13th Floor, 46021, Valencia, Spain

Telephone: +34 961 24 32 25

# Abstract

**Objective**

Automated machine learning (AutoML) frameworks can lower technical barriers for predictive and prognostic model development in radiomics by enabling researchers without programming expertise to build models. However, their effectiveness in addressing radiomics-specific challenges remains unclear. This study evaluates the performance, efficiency, and accessibility of general-purpose and radiomics-specific AutoML frameworks on diverse radiomics classification tasks, thereby guiding researchers and highlighting development needs for radiomics.

**Material and Methods**

Ten public/private radiomics datasets with varied imaging modalities (CT/MRI), sizes, anatomies and endpoints were used. Six general-purpose and five radiomics-specific frameworks were tested with predefined parameters using standardized cross-validation. Evaluation metrics included AUC, runtime, together with qualitative aspects related to software status, accessibility, and interpretability.

**Results**

Simplatab, a radiomics-specific tool with a no-code interface, achieved the highest average test AUC (81.81%) with a moderate runtime (~1 hour). LightAutoML, a general-purpose framework, showed the fastest execution with competitive performance (78.74% mean AUC in six minutes). Most radiomics-specific frameworks were excluded from the performance analysis due to obsolescence, extensive programming requirements, or computational inefficiency. Conversely, general-purpose frameworks demonstrated higher accessibility and ease of implementation.

**Conclusion**

Simplatab provides an effective balance of performance, efficiency, and accessibility for radiomics classification problems. However, significant gaps remain, including the lack of accessible survival analysis support and the limited integration of feature reproducibility and harmonization within current AutoML frameworks. Future research should focus on adapting AutoML solutions to better address these radiomics-specific challenges.

## Key points & Clinical Relevance

**Question:** Which open-source automated machine learning (AutoML) framework delivers the best performance and accessibility for radiomics-based problems across multiple organs and modalities?

**Findings:** Across ten heterogeneous radiomics datasets, the no-code Simplatab framework achieved the top average performance score (area under curve 81.8 %) with a moderate one-hour runtime.

**Clinical relevance:** Selecting a high-performing, user-friendly AutoML framework can speed translation of radiomics into routine imaging practice by lowering technical barriers and enabling reproducible model development.

## Acknowledgments

**Funding:** The authors declare that no funds, grants, or other support were received during the preparation of this manuscript.

## Abbreviations

| | |
|---|---|
| AutoML | Automated machine learning |
| WORC | Workflow for optimal radiomics classification |
| IBSI | Image biomarker standardization initiative |
| AUC | Area under the receiver operating characteristic curve |
| SD | Standard deviation |
| CI | Confidence interval |

## Keywords

# Introduction

In recent years, automated machine learning (AutoML) has emerged as a powerful approach to reduce technical barriers in machine learning model development. AutoML refers to the automation of the end-to-end machine learning workflow, including data preprocessing, feature selection, model selection, and hyperparameter optimization [1]. By delegating these technically demanding steps to algorithm-driven processes, AutoML allows clinicians and researchers without extensive machine learning experience to use advanced modeling techniques effectively, enabling medical professionals to concentrate on clinical questions and data interpretation instead of technical model building [2]. Modern AutoML frameworks promise improved model performance while promoting consistency and reproducibility by following structured, algorithm-driven processes that reduce individual variations. Similar to the introduction of genomics into clinical practice, medical professionals need transparent and interpretable models to understand radiomics features and apply them across the full spectrum of clinical endpoints, including classification tasks (e.g., tumor grading, diagnosis), regression problems (e.g., predicting continuous biomarkers or treatment response), and survival or time-to-event analyses that are essential in oncology research.

Radiomics represents a particularly challenging use case for AutoML in medical imaging. It involves the extraction of a large number of quantitative features from imaging exams for tissue characterization to serve as imaging biomarkers correlated with disease diagnosis or clinical endpoints [3]. The classical radiomics pipeline involves multiple sequential steps requiring domain-specific decisions and multidisciplinary expertise [4] (Figure 1), but faces significant limitations including lack of reproducibility and standardization, where small variations in image acquisition or processing substantially affect extracted features [5, 6]. Critical challenges include the need for harmonization methods to ensure feature consistency across different imaging platforms and institutions [7], and the high dimensionality of radiomics data relative to sample sizes, which creates highly correlated, sparse feature sets prone to overfitting that reduces feature robustness and biological interpretability across studies [8]. Numerous studies have shown the potential of radiomics features and radiomics-based models for cancer detection, tumor grading, and even predicting survival or treatment response, ultimately enhancing clinical decision-making [9–11]. Despite its potential, scaling radiomics into routine clinical practice remains challenging, and to date, no diagnostic tests or companion diagnostics based on radiomics have been successfully implemented.

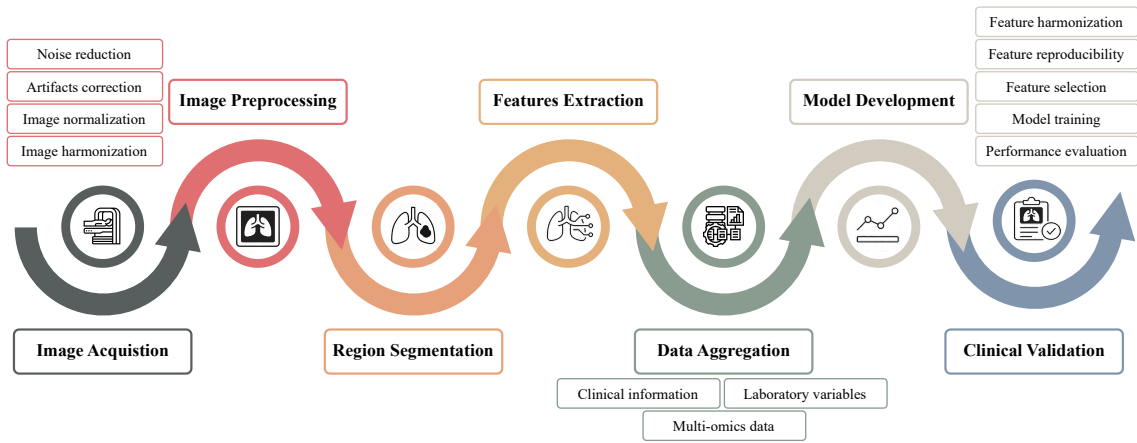

Figure 1. Overview of the standard radiomics pipeline. Image acquisition initiates the radiomics pipeline followed by preprocessing and the segmentation of the region of interest. Quantitative features are extracted, creating high-dimensional datasets. Prior to modeling, features may be harmonized to correct non-biological variations caused by differences in acquisition conditions, filtered for stability, or reduced. Features can be combined with additional data before training machine learning models to eventually achieve clinical validation.

To address these challenges, several radiomics-specific AutoML frameworks have been developed to explicitly address the complexity of radiomics workflows, automating steps from feature extraction to model building [12, 13]. In parallel, general-purpose AutoML, designed primarily for automating machine learning tasks on tabular data, have also been increasingly applied to radiomics studies. However, these efforts have largely remained method-specific, as most published studies have evaluated a single approach or a limited subset of tools in isolation [12–14]. Likewise, applications of general AutoML libraries to radiomics are often bespoke and tailored to specific use cases, without a unified methodological perspective across paradigms. [11]. As a result, current evidence provides only a fragmented view of how AutoML strategies behave in radiomics, particularly with respect to robustness, performance stability, and computational trade-offs across heterogeneous scenarios.

In this work, we explore the behavior of different AutoML frameworks in radiomics-based medical image classification. Using a unified experimental setup across ten heterogeneous radiomics datasets, we compare the performance and computational characteristics of both general-purpose and radiomics-oriented AutoML frameworks across diverse imaging modalities, anatomical regions, and clinical endpoints. Our ultimate goal is to provide methodological insights that help guide the design and practical deployment of AutoML solutions in radiomics, supporting more robust, reproducible, and clinically relevant modeling workflows.

# Materials and Methods

This retrospective computational evaluation was conducted in compliance with international ethical guidelines and regulations. For the eight publicly available datasets used, original study approvals apply. For the two private institutional datasets, the requirement for patient informed consent was waived by the corresponding institutional review board due to the retrospective and fully anonymized nature of the data.

## Datasets

The present study employed ten distinct radiomics datasets to evaluate the behavior of AutoML frameworks in radiomics-based classification tasks. These datasets included diverse imaging modalities, anatomical regions, varying class balances, and clinical endpoints, thereby offering comprehensive coverage of the radiomics applications (Table 1 and Supplementary Table 1). Eight publicly available datasets from OpenRadiomics (*BraTS, TCIA*) [15] and Workflow for Optimal Radiomics Classification (WORC; *CRLM, Desmoid, GIST, Lipo, Liver, Melanoma*) databases were included [16]. Additionally, two private institutional datasets were added to test the generalization in radiomics contexts not represented in the public domain (*Prostate* [17] and *Lung*). Patient cohorts ranged significantly in size from 74 to 577 subjects, reflecting variability between smaller single-institution datasets and larger multi-center collections. The inclusion and exclusion criteria for each cohort are detailed in their respective original publications [15–17].

All datasets provided pre-extracted radiomics features compliant with the Image Biomarker Standardization Initiative (IBSI). Therefore, the present evaluation focuses exclusively on the modeling and classification stages of the radiomics workflow, and not on the full end-to-end pipeline including image preprocessing, segmentation, and feature extraction.

Table 1. Summary of radiomics datasets included in the comparative analysis.

| Dataset | Accessibility | Size [n] | Image modality | Region of interest | Radiomic features | | Clinical variables [n] | Binary target |
|---|---|---|---|---|---|---|---|---|
| | | | | | Segmentation | Extracted [n] | | |
| *OpenRadiomics* | Public | | | | | | | |
|   *BraTS* | | 577 | MR-T1w | Brain: primary tumor | Manual | 1688 | - | MGMT: methylated / no methylated |
|   *TCIA* | | 421 | CT | NSCLC: primary tumor | Manual | 1688 | 3 (2 categorical) | TNM Overall Stage: I+II / III + IV |
| WORC | Public | | | | | | | |
|   *CRLM* | | 74 | CT | Colorectal: metastatic | Semi-automatic | 1379 | 2 (1 categorical) | HGP: Desmoplastic / Replacement |
|   *Desmoid* | | 202 | MR-T1w | Soft tissue: primary tumor | Semi-automatic | 1379 | 2 (1 categorical) | DTF / STS |
|   *GIST* | | 246 | CT | Gastrointestinal: lesion | Semi-automatic | 1379 | 2 (1 categorical) | GIST / No-GIST |
|   *Lipo* | | 114 | MR-T1w | Fat tissue: lesion | Semi-automatic | 1379 | 2 (1 categorical) | Lipoma / WDLPS |
|   *Liver* | | 185 | MR-T2w | Liver: primary tumor | Semi-automatic | 1379 | 2 (1 categorical) | Malign / Benign |
|   *Melanoma* | | 102 | CT | Lung: metastatic nodules | Semi-automatic | 1379 | 2 (1 categorical) | BRAF: mutated / wild type |
| *Lung* | Private | 554 | CT | NSCLC: primary tumor and lymph nodes | Semi-automatic | 1379 | 36 (2 categorical) | Survival at 12 months |
| *Prostate* | Private | 333 | MR-T2w | Prostate: central and transition zones | Semi-automatic | 1379 | - | Gleason: < 7 / ≥ 7 |

Abbreviations: WORC, Workflow for Optimal Radiomics Classification; NSLC, Non-Small Cell Lung Cancer; MGMT, $O^6$-methylguanine-DNA methyltransferase; TNM, Tumor - Node - Metastasis staging; HGP, Histopathological Growth Patterns; DTF, Desmoid-Type Fibromatosis; STS, Soft-Tissue Sarcomas; GIST, Gastrointestinal Stromal Tumors; WDLPS, Well-Differentiated Liposarcoma; BRAF, v-raf murine sarcoma viral oncogene homolog B1 gene.

## AutoML Frameworks

AutoML frameworks were selected based on two primary criteria: 1) open-source availability with no licensing fees, and 2) prominence in current literature, defined here as recurrent use or discussion in recent peer-reviewed studies within the fields of medical imaging, radiomics, and automated machine learning. This selection aimed to represent both major AutoML paradigms, namely general-purpose frameworks originally developed for tabular data and domain-specific approaches tailored to radiomics. General-purpose AutoML methods included: Autogluon [18], H2O AutoML [19], LightAutoML [20], MLjar [21], PyCaret [22] and TPOT [23], while radiomics-specific tools included: AutoRadiomics [13], AutoML for Radiomics [24], AutoPrognosis [25], Simplatab[14] and WORC [12].

## Evaluation Criteria

### Frameworks assessment

First, three qualitative aspects were assessed: obsolescence, accessibility and explainability. Obsolescence was examined through repository activity and update frequency. Repository status was categorized as "Active" (contributions within the last six months), "Maintenance" (activity between six months and two years), or "Obsolete" (inactive for over two years), while updates frequency was specifically rated as "High" (monthly releases), "Moderate" (at least every six months), or "Low" (sporadic updates less than once a year). Accessibility determined the barrier to entry based on installation complexity and required user expertise. Deployment quality was rated "High" for standard Python packages (accessible via *pip*, *uv*, or *Docker*) with comprehensive documentation, scaling down to "Low" for complex builds with sparse guidance. Concurrently, the required learning curve was categorized from "Low" for libraries adhering to standard conventions like the *scikit-learn* API, to "Advanced", which demands a deep expertise of the framework. Finally, interpretability was evaluated for their ability to provide model-level explanations of predictions, distinguishing between "Advanced" integration of model-agnostic methods (e.g., SHAP, LIME), "Basic feature importance reporting", and "None". These evaluations were conducted by a data scientist with four years of experience developing radiomics-based predictive and prognostic algorithms. Frameworks deemed Obsolete or exhibiting low accessibility were excluded from subsequent quantitative analysis.

Then, frameworks were quantitatively assessed based on their predictive performance using the area under the receiver operating characteristic curve (AUC) and computational efficiency based on the execution times during experiments. These two metrics were jointly analyzed to characterize the trade-off between predictive performance and computational efficiency across frameworks. An efficiency baseline was subsequently defined based on the joint assessment of predictive performance and computational runtime. Autogluon and MLjar offered different predefined performance configurations (presets), that were also evaluated as independent frameworks. Figure 2 summarizes the complete evaluation workflow.

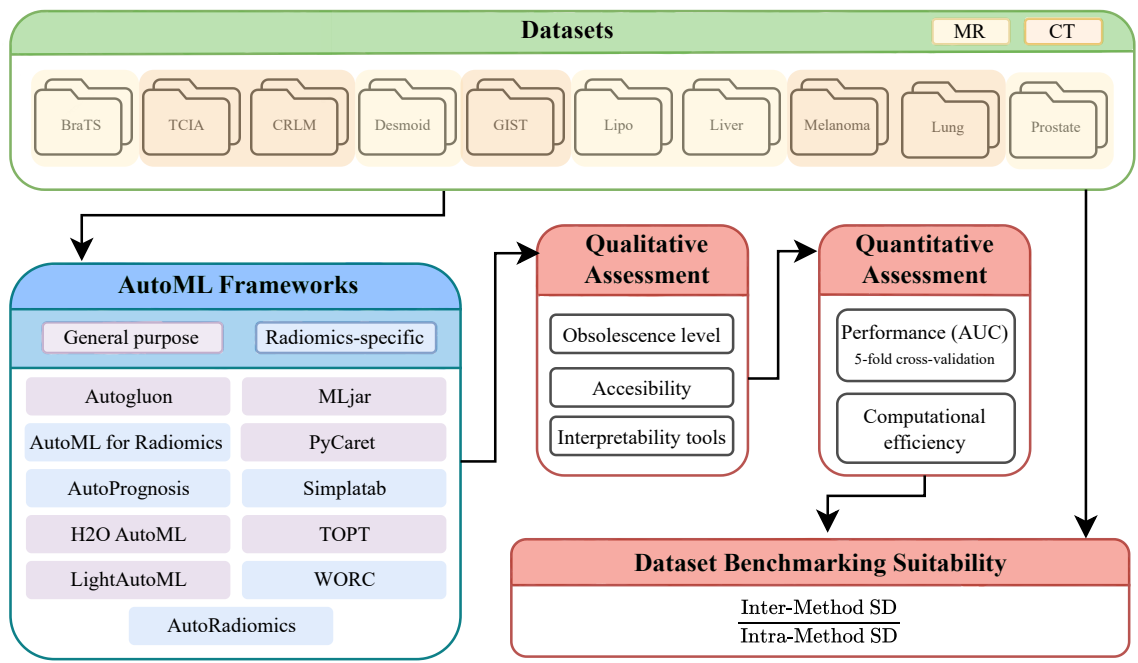

Figure 2. Methodological workflow for the comparative evaluation of AutoML frameworks. Frameworks were evaluated through qualitative and quantitative assessment while also evaluating dataset suitability for benchmarking. Obsolete or inaccessible frameworks were excluded from performance comparison.

To ensure methodological consistency, all datasets underwent minimal preprocessing to allow each AutoML method to apply its specific training pipeline. Exceptions to this procedure occurred only when certain frameworks had inherent limitations: median imputation for Simplatab (missing value incompatibility) and numerical encoding for TPOT (categorical variable incompatibility). Additionally, frameworks were initialized with predefined parameters and evaluated using 5-fold cross-validation under the same hardware with identical partitions across all methods. Performance was reported as mean AUC ± standard deviation (SD) across the five folds.

Additional details regarding radiomics feature extraction, framework pipelines and experimental setup are provided in the supplementary material. All the code is available in GitHub (https://github.com/joselznom/AutoML-Comparison-in-Radiomics/tree/main).

**Dataset suitability for AutoML evaluation in radiomics**

An effective dataset for comparative methodological evaluation should enable measuring consistent and discriminative signals of methodological differences. We adapted criteria from medical image segmentation validation to assess dataset suitability for AutoML in radiomics [26]. Two main requirements were defined to categorize a dataset as suitable for benchmarking: 1) low SD of AUC scores from the same method across the five folds (intra-method SD). This indicates statistical stability, ensuring that a method's observed performance is consistent and not due to random chance or specific data splits; and 2), high SD across different methods (inter-method SD) indicating the presence of meaningful signals of methodological differences. Low inter-method SD would suggest that the task is too simple or that performance saturates quickly, limiting its utility for distinguishing methodological superiority.

The final suitability score was defined as the ratio between inter-method and intra-method SD. This metric functions as a dimensionless signal-to-noise indicator of how clearly methodological differences can be detected. A ratio exceeding a parity threshold of 1.0 indicates that algorithmic superiority is distinct and distinguishable above the inherent noise. Datasets with low intra-method SD and high inter-method SD are preferred as they offer high differentiation power and result stability.

**Statistical analysis**

Statistical evaluation of performance differences was conducted using Python 3.11 (SciPy v.1.12). A two-sided paired Wilcoxon signed-rank test was applied to compare the top-performing framework (those with the highest mean AUC across datasets) and the efficiently baseline against each alternative. *P*-values were adjusted for multiple comparisons using the Benjamini-Hochberg false discovery rate correction, with corrected *p*-values < 0.05 considered statistically significant.

# Results

## Dataset suitability for comparative methodological evaluation

Figure 3 summarizes the overall suitability of the evaluated datasets for comparative methodological analysis, measured as the ratio of inter-method to intra-method AUC SD.

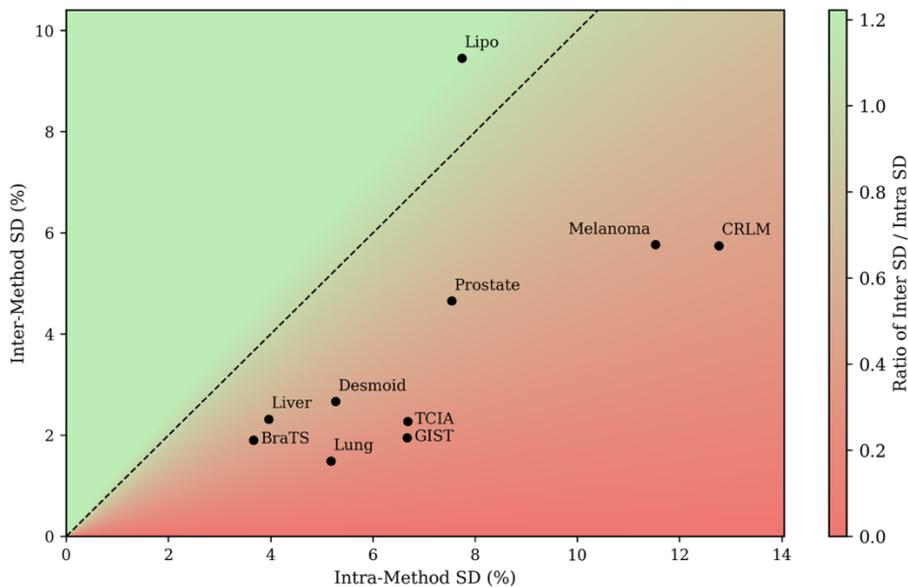

Figure 3. Dataset suitability for comparative methodological evaluation. Suitability was quantified as the ratio between inter-method and intra-method standard deviation (SD) based of the mean AUC obtained across AutoML frameworks. The horizontal axis (intra-method SD%) reflects the internal variability of AUC values within each method across cross-validation folds, whereas the vertical axis (inter-method SD%) represents the variability of AUC values across different methods. The dashed line denotes a ratio of one. Datasets located in the upper-left (green) region combine low internal variability with high inter-method differentiation. AUC, area under the curve.

As observed, datasets such as *Lipo* illustrated scenarios with a favorable balance between result stability and inter-method differentiation. Conversely, datasets like *CRLM* and *Melanoma*, exhibited high intra-method variability, limiting the reliability of performance comparisons. In between these two extremes, *Prostate* exhibited intermediate behavior, with relatively higher internal variability than the most stable datasets. Finally, *TCIA*, *GIST*, *Desmoid*, *Lung*, *Liver*, and *BraTS* demonstrated stable results (low intra-method SD) but limited discriminatory capacity (low inter-method SD) between methods.

## Qualitative Assessment

Table 2 provides a summary of the qualitative characteristics for all initially considered frameworks. The assessment revealed limited viability for several radiomics-specific tools. Thus, AutoML for Radiomics and AutoRadiomics were classified as obsolete, with inactive repositories and no recent updates, rendering them unusable in the experimental setup. WORC, while actively maintained, proved incompatible with the pre-extracted radiomics features used in this study and demanded a high level of programming expertise, deviating from the low-code paradigm typical of AutoML. AutoPrognosis, although functional and offering advanced capabilities, exhibited very high computational demands, failing to complete the first cross-validation fold within 72 hours on typical high-dimensional radiomics datasets, which led to its exclusion from the subsequent quantitative analysis. Consequently, only Simplatab remained as a viable radiomics-specific framework for the quantitative comparison.

In contrast, general-purpose frameworks such as LightAutoML, MLjar, and Autogluon demonstrated active maintenance, straightforward installation procedures, and lower requirements for machine learning expertise.

Notably, several frameworks, including MLjar, Simplatab, and TPOT, offered advanced interpretability tools (e.g., SHAP values and model bias analysis)—a key differentiator for clinical translation and trustworthiness—, whereas others provided only limited or no interpretability functionality.

## Quantitative Performance and Efficiency

The quantitative evaluation was performed on the general-purpose frameworks and Simplatab, following the qualitative filtering described above. Table 3 presents the detailed AUC results (mean ± SD) for each framework across the ten datasets.

Overall, Simplatab achieved the highest average AUC (81.81%) across datasets; a statistically superior performance compared to all other evaluated frameworks. In turn, LightAutoML delivered competitive results (78.20% AUC) and was therefore selected as the efficiency baseline, as it achieved a favorable balance between performance and computational time, with training completed within minutes. Although its performance was statistically lower than that of Simplatab's, it remained significantly higher than that of H2O AutoML and PyCaret. Conversely, H2O AutoML and PyCaret exhibited statistically significant underperformance, suggesting limited suitability for these radiomics classification tasks under the evaluated conditions. The remaining frameworks, including the different Autogluon and MLjar presets, showed no statistically significant difference in terms of performance when compared to LightAutoML, although they were consistently outperformed by Simplatab

Table 2. Qualitative characteristics of the general-purpose and radiomics-specific AutoML frameworks.

| Framework | Focus | Type of Problem | Obsolescence Level | | Accessibility | | Interpretability Tools [2] |
|---|---|---|---|---|---|---|---|
| | | | Repository Activity | Updates Frequency [1] | Ease of Installation | Required ML Knowledge | |
| Autogluon | General-purpose | Classification, regression, time series | Active | High | High | Low | None |
| AutoML for Radiomics | Radiomics | Classification | Obsolete | - | Not possible | Intermediate | Basic |
| AutoPrognosis | Healthcare | Classification, regression, survival | Active | High | High | Low | Advanced |
| AutoRadiomics | Radiomics | Classification | Obsolete | Low | Medium | Intermediate | Basic |
| H2O AutoML | General-purpose | Classification, regression | Active | High | High | Low | None |
| LightAutoML | General-purpose | Classification, regression | Active | High | High | Low | Basic |
| MLjar | General-purpose | Classification, regression | Active | High | High | Low | Advanced |
| PyCaret | General-purpose | Classification, regression, time series, clustering, anomaly detection | Active | Moderate | High | Low | None |
| Simplatab | Radiomics | Classification | Active | Moderate | High | Low | Advanced |
| TPOT | General-purpose | Classification, regression | Maintenance | Moderate | High | Low | None |
| WORC | Radiomics | Classification | Maintenance | Moderate | Not possible | Advanced | None |

[1] High: one or more times per month, Moderate: at least every six months, Low: less than once a year.
[2] Basic: offers basic visualization or basic information about the features of the model, Advanced: includes advanced techniques like SHAP, LIME, heatmap visualization, etc.
Abbreviations: WORC, Workflow for Optimal Radiomics Classification; ML, machine learning.

Table 3. Results of the comparative evaluation of AutoML frameworks measured as AUC (%) reported as mean ± standard deviation (SD). Statistical comparisons are indicated. Color intensity increases with higher scores and is normalized per column. The "Average" column shows the average AUC across datasets for each framework, excluding *CRLM* and *Melanoma* due to the lack of reliable performance across frameworks.

| | BraTS n = 577 | CRLM n = 74 | Desmoid n = 202 | Lipo n = 114 | Liver n = 185 | Melanoma n = 102 | Prostate n = 333 | TCIA n = 421 | Lung n = 554 | GIST n = 246 | Average | Runtime |
|---|---|---|---|---|---|---|---|---|---|---|---|---|
| Autogluon – Medium | 58.7 ± 3.5 | 63.2 ± 18.0 | 89.7 ± 8.7 | 85.6 ± 6.1 | 92.5 ± 4.4 | 56.5 ± 22.0 | 68.8 ± 9.5 | 67.7 ± 4.6 | 79.4 ± 4.9 | 75.3 ± 5.3 | 77.21 [*] | 3 m |
| Autogluon – Good | 60.5 ± 4.5 | 53.1 ± 17.1 | 93.2 ± 3.9 | 81.1 ± 13.2 | 94.9 ± 2.8 | 51.1 ± 13.2 | 67.4 ± 10.6 | 70.0 ± 7.0 | 80.3 ± 5.9 | 78.7 ± 5.9 | 78.26 [*] | 1.5 h |
| Autogluon – High | 58.9 ± 5.5 | 50.8 ± 9.2 | 92.4 ± 4.1 | 85.5 ± 9.9 | 94.1 ± 3.4 | 50.2 ± 12.9 | 68.1 ± 9.4 | 71.2 ± 6.8 | 80.0 ± 5.9 | 79.7 ± 5.0 | 78.74 [*] | 5 h |
| Autogluon – Best | 57.7 ± 3.6 | 57.3 ± 8.8 | 93.1 ± 4.0 | 80.2 ± 11.8 | 95.0 ± 3.3 | 52.3 ± 12.9 | 70.2 ± 9.1 | 70.2 ± 7.3 | 80.9 ± 5.2 | 79.7 ± 6.6 | 78.38 [*] | 5 h |
| MLjar – Explain | 58.6 ± 4.0 | 56.3 ± 11.3 | 92.0 ± 5.4 | 83.3 ± 8.6 | 88.6 ± 5.0 | 51.4 ± 14.3 | 63.9 ± 9.3 | 70.0 ± 8.3 | 77.2 ± 2.9 | 75.1 ± 7.1 | 76.09 [*] | 17 m |
| MLjar – Perform | 59.5 ± 3.4 | 58.4 ± 11.6 | 91.1 ± 5.5 | 81.4 ± 8.3 | 94.2 ± 2.6 | 52.9 ± 14.6 | 68.7 ± 11.3 | 71.6 ± 7.1 | 81.1 ± 5.1 | 78.8 ± 7.6 | 78.30 [*] | 5 h |
| MLjar – Compete | 59.7 ± 2.2 | 56.0 ± 20.3 | 88.6 ± 4.0 | 76.8 ± 11.5 | 90.0 ± 5.5 | 51.6 ± 7.6 | 70.3 ± 7.9 | 66.9 ± 6.2 | 78.3 ± 6.1 | 78.5 ± 7.1 | 76.14 [*] | 5 h |
| MLjar – Optuna | 60.5 ± 3.7 | 53.0 ± 18.1 | 90.7 ± 5.6 | 81.7 ± 8.3 | 93.9 ± 3.3 | 43.3 ± 9.5 | 72.5 ± 4.4 | 70.5 ± 4.5 | 80.6 ± 3.4 | 78.4 ± 6.6 | 78.60 [*] | 34 h |
| H2O AutoML | 56.9 ± 4.3 | 50.0 ± 0.0 | 85.8 ± 7.2 | 50.0 ± 0.0 | 89.4 ± 6.3 | 50.0 ± 0.0 | 68.2 ± 5.9 | 70.9 ± 5.3 | 78.2 ± 6.8 | 76.5 ± 8.0 | 71.99 [*▼] | 30 s |
| LightAutoML | 56.7 ± 4.2 | 52.3 ± 14.2 | 92.6 ± 4.4 | 84.4 ± 3.2 | 93.5 ± 4.6 | 41.7 ± 9.8 | 68.9 ± 3.0 | 71.5 ± 5.3 | 80.8 ± 5.0 | 77.2 ± 7.2 | 78.20 [*] | 6 m |
| PyCaret | 56.2 ± 3.0 | 42.8 ± 12.4 | 86.5 ± 9.6 | 81.6 ± 6.1 | 93.0 ± 4.3 | 52.2 ± 15.0 | 54.5 ± 4.7 | 65.7 ± 9.4 | 78.8 ± 5.2 | 76.8 ± 8.6 | 74.14 [*▼] | 16 m |
| Simplatab [1] | **63.3 ± 2.4** | **64.5 ± 10.0** | **95.0 ± 3.8** | **87.7 ± 5.5** | **96.4 ± 2.3** | **65.6 ± 7.5** | **73.3 ± 5.8** | **74.1 ± 7.1** | **82.4 ± 5.9** | **82.3 ± 4.4** | **81.81** [▲] | ~1 h |
| TPOT | 58.4 ± 3.3 | 58.6 ± 15.0 | 91.0 ± 2.3 | 78.2 ± 8.1 | 92.7 ± 3.7 | 49.2 ± 10.5 | 67.3 ± 7.1 | 67.8 ± 7.9 | 81.4 ± 5.0 | 78.2 ± 7.3 | 76.88 [*] | 2.5 h |

[1] Runtime for Simplatab is approximate, as the tool works with a web interface, making direct time comparison was not possible.

[*] Statistically significant lower performance compared to the top-performer Simplatab (paired Wilcoxon signed-rank test with Benjamini-Hochberg correction, $p < 0.05$).

▲/▼ Statistically significant higher/lower performance compared to the efficiency baseline LightAutoML (paired Wilcoxon signed-rank test with Benjamini-Hochberg corrections, $p < 0.05$).

In line with the suitability analysis (Figure 3), *CRLM* and *Melanoma* yielded poor performance across most frameworks (AUC ≤ 60% with high SD), reflecting the intrinsic instability of these datasets. Nevertheless, Simplatab showed relatively greater robustness on these challenging scenarios, achieving modestly higher AUC values (around 65%) with lower variance, which suggests improved tolerance to noisy and unstable data. In contrast, framework performance was relatively homogeneous in low-differentiation datasets, confirming their limited utility for methodological discrimination. For instance, AUC differences across frameworks in the *Lung* and *TCIA* datasets were minimal, further corroborating the findings derived from Figure 3.

Figure 4 presents the relationship between average AUC and runtime for each AutoML framework. From this joint perspective, a clear efficiency frontier emerges, along which frameworks such as Simplatab and LightAutoML attain high AUC scores with substantially reduced runtimes, indicating a favorable performance-efficiency trade-off. The remaining frameworks clearly underperformed this joint criterion. Among these secondary options, only Autogluon presets remained partially competitive (i.e., *Medium Quality* vs. LightAutoML and *High Quality* vs. Simplatab) although generally at the cost of slower or slightly reduced performance.

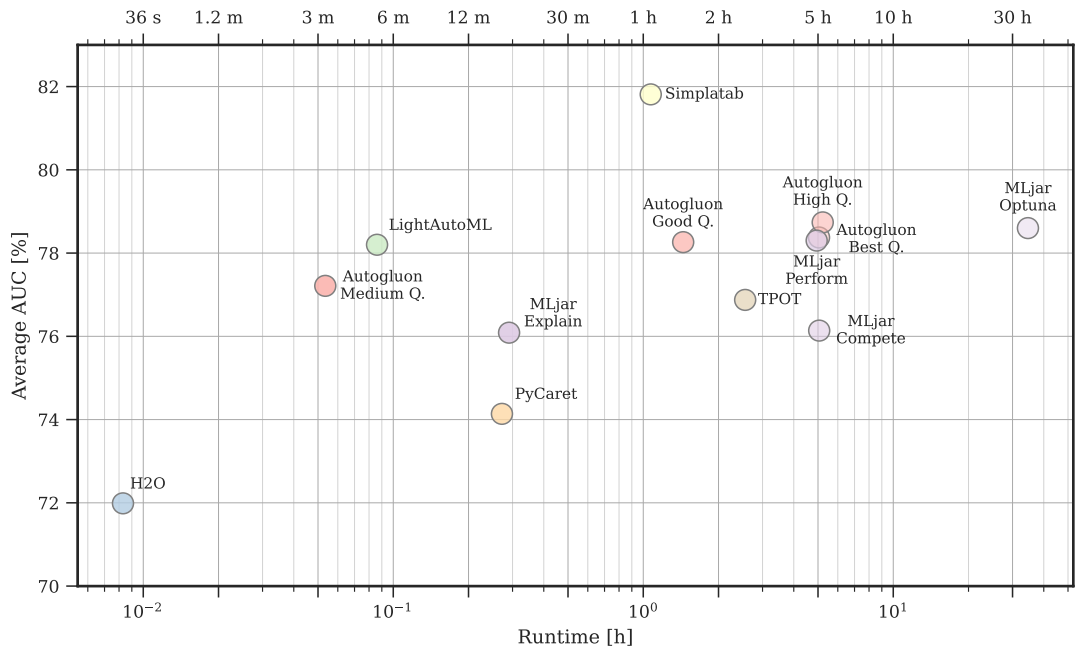

Figure 4: Performance and computational efficiency trade-offs of the evaluated AutoML frameworks. Test average AUC [%] across all datasets is plotted against total runtime [h] on a logarithmic scale. The upper axis displays representative runtime values in natural scale to aid interpretation. Different presets of Autogluon and MLjar are shown separately. AUC, area under the curve.

# Discussion

The comparative analysis of general-purpose and radiomics-specific AutoML frameworks for radiomics-based classification tasks reveals significant heterogeneity in performance, efficiency, computational costs, accessibility, and technical maturity, providing practical insights into the current capabilities of AutoML for radiomics applications.

A primary finding was Simplatab's statistically superior performance (average AUC 81.81%), computational efficiency (~1h runtime), and relative robustness on challenging datasets such as *CRLM* and *Melanoma*, which exhibited high intra-method SD and poor performance across most other frameworks. Beyond performance, Simplatab's key differentiator lied in its accessibility and its suite of tools aimed at facilitating model inspection. The framework provides a graphical user interface that requires no coding expertise, together with built-in modules for model interpretability and basic assessments of data imbalance and model behavior. While these tools do not provide causal interpretability, they represent practical mechanisms to increase model transparency and support user trust during model exploration and validation that may facilitate the gradual integration of radiomics-based models into clinical research workflows.

In contrast, general-purpose AutoML frameworks showed varying strengths and weaknesses. Autogluon (*High Quality* preset) and MLjar (*Optuna* preset) delivered competitive predictive performance, but often with longer runtimes. LightAutoML emerged as an exceptionally efficient option, providing good AUC scores in minutes with no statistically significant performance difference relative to other frameworks requiring substantially longer execution time, positioning it as a suitable tool for rapid prototyping and preliminary analyses. Conversely, H2O, despite being very fast, yielded lower predictive performance under default configurations, suggesting that its standard pipelines may be less suited for the complexities often inherent in high-dimensional radiomics datasets.

Our qualitative assessment highlighted significant limitations among several radiomics-specific tools. Frameworks such as AutoRadiomics and AutoML for Radiomics, appeared to be inactive or technologically obsolete, reflecting the challenge of maintaining specialized software in a rapidly evolving ML landscape. WORC, while actively maintained, relied on configuration files and scripting that demand considerable programming expertise, which conflicts with the low-code/no-code paradigm typically associated with AutoML solutions. AutoPrognosis, despite its advanced methodological scope, proved computationally infeasible for the high-dimensional datasets considered in this study under the adopted experimental constraints. Consequently, Simplatab was the only radiomics-specific framework suitable for inclusion in the quantitative analysis.

The dataset suitability analysis revealed limitations in several public radiomics datasets: *CRLM* and *Melanoma* showed high intra-method variability, making it difficult to discern true performance differences from statistical noise, whereas others (*Lung*, *TCIA*, and *Liver*) showed low inter-method variability, which limits their ability to discriminate between modeling pipelines. These findings highlight the need for larger, more diverse, and statistically stable radiomics datasets to support meaningful methodological comparisons and advancements.

Our study has limitations that highlight critical gaps in the current AutoML landscape for radiomics. First, the present evaluation focused exclusively on classification tasks. The absence of robust, user-friendly AutoML frameworks supporting survival analysis represents a major bottleneck for oncology, where time-to-event endpoints are paramount. Future development should prioritize the integration of survival models into accessible AutoML packages. Second, the use of pre-extracted features isolated the modeling component of the radiomics pipeline but does not address the ultimate goal of achieving a fully automated, end-to-end workflow. A major unmet need is the development of frameworks that also integrate upstream steps such as feature extraction, harmonization and reproducibility control, which remain critical barriers for clinical translation. While tools like Simplatab demonstrate promise at the modeling stage, to ensure reliability and clinical trust, the field should progress towards holistic solutions that automate the entire radiomics workflow, from image to prediction.

In conclusion, AutoML frameworks offer substantial potential to accelerate and democratize radiomics research by automating complex modeling tasks. Simplatab stands out as a particularly promising tool for users prioritizing ease of use, model inspection, and strong predictive performance with reasonable computational cost. General-purpose tools such as LightAutoML provide efficient options for rapid experimentation. However, substantial challenges remain. The field requires more stable and diverse radiomics datasets, AutoML solutions capable of supporting survival analysis, and, critically, frameworks that enable robust harmonization and reproducibility across the full radiomics workflow. Although tools such as Simplatab represent a significant step forward for the modeling stage, considerable development is still required to achieve fully automated, reliable, and clinically translatable radiomics pipelines.

# References

[1] G. A. Castro *et al.*, "Automated Machine Learning in medical research: A systematic literature mapping study," *Artif Intell Med*, vol. 171, p. 103302, Jan. 2026, doi: 10.1016/j.artmed.2025.103302.

[2] A. J. Thirunavukarasu *et al.*, "Clinical performance of automated machine learning: a systematic review," Oct. 26, 2023, *medRxiv*. doi: 10.1101/2023.10.26.23297599.

[3] "Radiomics: Images Are More than Pictures, They Are Data | Radiology." Accessed: May 06, 2024. [Online]. Available: https://pubs.rsna.org/doi/10.1148/radiol.2015151169

[4] J. E. van Timmeren, D. Cester, S. Tanadini-Lang, H. Alkadhi, and B. Baessler, "Radiomics in medical imaging—'how-to' guide and critical reflection," *Insights Imaging*, vol. 11, no. 1, p. 91, Dec. 2020, doi: 10.1186/s13244-020-00887-2.

[5] D. Veiga-Canuto *et al.*, "Reproducibility Analysis of Radiomic Features on T2-weighted MR Images after Processing and Segmentation Alterations in Neuroblastoma Tumors," *Radiol Artif Intell*, vol. 6, no. 4, p. e230208, July 2024, doi: 10.1148/ryai.230208.

[6] A. Traverso, L. Wee, A. Dekker, and R. Gillies, "Repeatability and Reproducibility of Radiomic Features: A Systematic Review," *Int J Radiat Oncol Biol Phys*, vol. 102, no. 4, pp. 1143–1158, Nov. 2018, doi: 10.1016/j.ijrobp.2018.05.053.

[7] J. Lozano-Montoya and A. Jimenez-Pastor, "Harmonization in the Features Domain," in *Basics of Image Processing: The Facts and Challenges of Data Harmonization to Improve Radiomics Reproducibility*, Á. Alberich-Bayarri and F. Bellvís-Bataller, Eds., Cham: Springer International Publishing, 2023, pp. 145–166. doi: 10.1007/978-3-031-48446-9_7.

[8] A. Demircioğlu, "Reproducibility and interpretability in radiomics: a critical assessment," *dir*, Oct. 2024, doi: 10.4274/dir.2024.242719.

[9] I. Fornacon-Wood, C. Faivre-Finn, J. P. B. O'Connor, and G. J. Price, "Radiomics as a personalized medicine tool in lung cancer: Separating the hope from the hype," *Lung Cancer*, vol. 146, pp. 197–208, Aug. 2020, doi: 10.1016/j.lungcan.2020.05.028.

[10] J. Lozano-Montoya *et al.*, "Risk stratification in neuroblastoma patients through machine learning in the multicenter PRIMAGE cohort," *Front. Oncol.*, vol. 15, Feb. 2025, doi: 10.3389/fonc.2025.1528836.

[11] X. Su *et al.*, "Automated machine learning based on radiomics features predicts H3 K27M mutation in midline gliomas of the brain," *Neuro Oncol*, vol. 22, no. 3, pp. 393–401, Mar. 2020, doi: 10.1093/neuonc/noz184.

[12] M. P. A. Starmans *et al.*, "An automated machine learning framework to optimize radiomics model construction validated on twelve clinical applications," Mar. 10, 2025, *arXiv*: arXiv:2108.08618. doi: 10.48550/arXiv.2108.08618.

[13] P. Woznicki, F. Laqua, T. Bley, and B. Baeßler, "AutoRadiomics: A Framework for Reproducible Radiomics Research," *Front Radiol*, vol. 2, p. 919133, July 2022, doi: 10.3389/fradi.2022.919133.

[14] D. I. Zaridis *et al.*, "Simplatab: An Automated Machine Learning Framework for Radiomics-Based Bi-Parametric MRI Detection of Clinically Significant Prostate Cancer," *Bioengineering*, vol. 12, no. 3, Art. no. 3, Mar. 2025, doi: 10.3390/bioengineering12030242.

[15] K. Namdar, M. W. Wagner, B. B. Ertl-Wagner, and F. Khalvati, "Open-radiomics: A Collection of Standardized Datasets and a Technical Protocol for Reproducible Radiomics Machine Learning Pipelines," Oct. 24, 2023, *arXiv*: arXiv:2207.14776. doi: 10.48550/arXiv.2207.14776.

[16] M. P. A. Starmans *et al.*, "The WORC database: MRI and CT scans, segmentations, and clinical labels for 930 patients from six radiomics studies," Aug. 25, 2021, *medRxiv*. doi: 10.1101/2021.08.19.21262238.

[17] H Kondylakis *et al.*, "ProCAncer-I: A platform integrating imaging data and AI models, supporting precision care through prostate cancer's continuum," presented at the IEEE-EMBS International Conference on Biomedical and Health Informatics (BHI), Loannina, Greece, Sept. 2022. doi: 10.3030/952159.

[18] N. Erickson *et al.*, "AutoGluon-Tabular: Robust and Accurate AutoML for Structured Data," Mar. 13, 2020, *arXiv*: arXiv:2003.06505. doi: 10.48550/arXiv.2003.06505.

[19] E. LeDell and S. Poirier, "H2O AutoML: Scalable Automatic Machine Learning".

[20] A. Vakhrushev, A. Ryzhkov, M. Savchenko, D. Simakov, R. Damdinov, and A. Tuzhilin, "LightAutoML: AutoML Solution for a Large Financial Services Ecosystem," Apr. 05, 2022, *arXiv*: arXiv:2109.01528. doi: 10.48550/arXiv.2109.01528.

[21] Płońska, Aleksandra and Płoński, Piotr, *MLJAR: State-of-the-art Automated Machine Learning Framework for Tabular Data*. (Apr. 01, 2021). Python. MLJAR, Łapy, Poland. Accessed: Apr. 01, 2025. [Online]. Available: https://github.com/mljar/mljar-supervised

[22] Moez, Ali, *PyCaret: An open source, low-code machine learning library in Python*. (Apr. 01, 2020). Accessed: Apr. 01, 2025. [Online]. Available: https://www.pycaret.org

[23] R. S. Olson and J. H. Moore, "TPOT: A Tree-Based Pipeline Optimization Tool for Automating Machine Learning," in *Automated Machine Learning*, F. Hutter, L. Kotthoff,

and J. Vanschoren, Eds., in The Springer Series on Challenges in Machine Learning. , Cham: Springer International Publishing, 2019, pp. 151–160. doi: 10.1007/978-3-030-05318-5_8.

[24] *Auto-ML for Radiomics Analysis*. (Feb. 15, 2025). Python. TAILab (Translational Artificial Intelligence Laboratory). Accessed: Apr. 01, 2025. [Online]. Available: https://github.com/Yonsei-TAIL/Auto-ML

[25] F. Imrie, B. Cebere, E. F. McKinney, and M. van der Schaar, "AutoPrognosis 2.0: Democratizing diagnostic and prognostic modeling in healthcare with automated machine learning," *PLOS Digit Health*, vol. 2, no. 6, p. e0000276, June 2023, doi: 10.1371/journal.pdig.0000276.

[26] F. Isensee *et al.*, "nnU-Net Revisited: A Call for Rigorous Validation in 3D Medical Image Segmentation," July 25, 2024, *arXiv*: arXiv:2404.09556. doi: 10.48550/arXiv.2404.09556.

# Supplementary Material

## Radiomics Feature Extraction

Radiomic feature extraction was performed differently depending on the dataset's origin. For the publicly available datasets obtained from OpenRadiomics (*BraTS* and *TCIA*), pre-computed radiomics feature sets were utilized. Both feature sets were originally generated using a *binwidth* parameter of 25 with z-score normalization.

For the remaining evaluated datasets (C*RLM, Desmoid, GIST, Lipo, Liver, Melanoma, Lung, Prostate*), radiomics features were extracted using the PyRadiomics python library (v3.1.0) [1]. A standardized configuration pipeline was implemented in PyRadiomics, consisting of image resampling to isotropic voxel, z-score normalization to standardize intensity values, and removal of outliers deviating by more than three standard deviations from the mean within the region of interest to mitigate their potential influence.

Feature extraction included shape, first- and second-order features from the original image, along and high-order features derived from filtered image representations. The applied filters included: square, exponential, logarithm, wavelet transforms, and Laplacian of gaussian with sigma values specified as 0.5, 3.0, and 5.0. This extraction protocol yielded a total of 1379 distinct radiomic features per case.

## Experimental Setup

Experiments were executed on a computing environment comprising an Intel Xeon CPU E3-1240 v6 (4 cores at 3.70 GHz), 33 GB of RAM, running CentOS Linux 7. Python versions 3.8 or 3.10 were used, depending on compatibility requirements with the specific frameworks.

The reliance on a CPU-based environment for all experiments was intentional and reflects the nature of state-of-the-art AutoML for tabular data. The leading algorithms for tabular data, such as the gradient-boosted trees and ensembles that power most frameworks, are highly optimized for efficient CPU execution. While GPU support exists for some of these algorithms, the performance benefits are typically only realized on datasets far larger than those common in radiomics.

# Datasets

## Public datasets sources

- OpenRadiomics datasets:
    - BraTS: https://openradiomics.org/?page_id=1163
    - TCIA: https://openradiomics.org/?page_id=1144
- WORC datasets: https://xnat.health-ri.nl/data/projects/worc

## Variable response distribution

Supp. Table 1: Summary of response variable distributions for each dataset.

| Dataset | Response Variable Description | Distribution [n] | |
|---|---|---|---|
| BraTS | MGMT | Non-methylated = 276<br>Methylated = 301 | |
| TCIA | TNM Overall Stage | I = 93<br>II = 40<br>IIIa = 112<br>IIIb = 176 | *Binarized:*<br>Stage I + II = 133<br>Stage III = 288 |
| CRLM | Histological Growth Pattern (HGP) | Desmoplastic HGP = 38<br>Replacement HGP = 36 | |
| Desmoid | Tumor Type | Soft Tissue Sarcoma (STS) = 131<br>Desmoid-Type Fibromatosis (DTF) = 71 | |
| GIST | Tumor Type | Non-GIST = 121<br>Gastrointestinal Stromal Tumor (GIST) = 125 | |
| Lipo | Tumor Type | Lipoma = 57<br>Well-Differentiated Liposarcoma (WDLPS) = 57 | |
| Liver | Lesion Malignancy | Malignant = 91<br>Bening = 94 | |
| Melanoma | BRAF Mutation Status | BRAF Mutated = 50<br>BRAF Wild Type = 52 | |
| Lung | Survival Status at 12 Months | Survived = 260<br>Deceased = 294 | |
| Prostate | Gleason Score | 6 = 78<br>7 = 200<br>8 = 24<br>9 = 28<br>10 = 2 | *Binarized:*<br>< 7 = 78<br>≥ 7 = 252 |

Abbreviations: MGMT, O$^6$-methylguanine-DNA methyltransferase; TNM, Tumor - Node - Metastasis staging; BRAF, v-raf murine sarcoma viral oncogene homolog B1 gene.

# AutoML Frameworks Pipelines

Supp. Table 2a: Comparative summary of the AutoML frameworks evaluated in the performance comparison based on feature selection strategies, supported machine learning models, and additional functionalities

| Framework | Feature Selection | Machine Learning Models | Additional Functionalities |
|---|---|---|---|
| Autogluon (v.1.1.0) | No advanced feature selection by default; removes constant and duplicate features. | LightGBM, XGBoost, CatBoost, Random Forest, KNN, logistic regression, PyTorch and FastAI neural networks; uses ensemble methods (stacking). | Preset configurations balance accuracy and speed with hyperparameter search: <br> - *Medium:* default preset for fastest training/inference and lowest accuracy. <br> - *Good:* fast with good accuracy (~4× faster than *High*). <br> - *High:* strong accuracy with moderate speed. <br> - *Best:* maximum accuracy, slowest training/inference. |
| MLjar (v.1.1.9) | - *Explain:* No feature selection. <br> - *Others preset:* Automatic iterative feature selection using permutation-based importance with random feature comparison across multiple learners. | Linear/logistic regression, decision trees, Random Forest, ExtraTrees, LightGBM, XGBoost, CatBoost, neural networks, KNN; combines top performers using ensemble methods and hill-climbing optimization. | - *Explain:* maximum interpretability (learning curves, importance plots, and SHAP plots) with minimal computation (minimal tuning with only 1 model per algorithm using default hyperparameters). Fastest option with basic 75%/25% train/test split. <br> - *Perform:* moderate explanations (learning curves and importance plots only) with balanced tuning (13 total models, including some hyperparameter optimization). 5-fold cross-validation for better performance estimation <br> - *Compete:* No explanations generated with extensive tuning (22 total models and aggressive hyperparameter search) 10-fold cross-validation for most robust evaluation with advanced ensemble methods (stacking and ensemble stacked). <br> - *Optuna*: *compete* preset with intensive tuning using Optuna. |

Abbreviations: CV, Cross-Validation; DNN, Deep Neural Network; GLM, Generalized Linear Model; KNN, K-Nearest Neighbors, MLP, Multilayer Perceptron; RFE, Recursive Feature Elimination; SGD; Stochastic Gradient Descent; SHAP, Shapley Additive exPlanations; SVM, Support Vector Machine, XRT, eXtra Random Trees.

Supp. Table 2b: Comparative summary of the AutoML frameworks evaluated in the performance comparison based on feature selection strategies, supported machine learning models, and additional functionalities.

| Framework | Feature Selection | Machine Learning Models | Additional Functionalities |
|---|---|---|---|
| H2O (v.3.46.0) | No additional feature selection; performs automatic target encoding for high-cardinality categorical variables. | Multiple H2O-native algorithms: GLM with regularization, XGBoost, LightGBM, Random Forest, DNN, XRT; produces stacked ensembles. | Parallel model execution with early stopping; produces ensembles with regularized meta-learner. |
| LightAutoML (v.0.4.1) | Feature preselection based on importance and sequential selection | Multi-level pipelines with LightGBM, CatBoost, Random Forest; uses stacking with weighted blending. | Automatic hyperparameter tuning with Optuna and cross-validation. |
| PyCaret (v.3.3.2) | Feature selection through SelectFromModel (default: LightGBM), SelectKBest, or SequentialFeatureSelector; includes multicollinearity removal. | Wide range of scikit-learn and boosting algorithms: linear/logistic regression, KNN, SVM, Naive Bayes, decision trees, Random Forest, ExtraTrees, AdaBoost, Gradient Boosting, LightGBM, CatBoost, XGBoost, MLP; supports basic ensembling. | Hyperparameter tuning with random search and cross-validation. |
| Simplatab (v.1.0.0) | Removes highly correlated features, then applies combined approach: SULOV (stability selection) + RFE using XGBoost as base model. | Seven classifiers: logistic regression, decision trees, Random Forest, XGBoost, multilayer perceptron (MLP), SGD classifier, SVM; hyperparameters optimized with cross-validation. | Includes bias detection and model vulnerability analysis. |
| TPOT (v.1.1.0) | Genetic feature selection during evolutionary pipeline search, incorporating filtering nodes. | Scikit-learn estimators: trees, SVM, KNN, linear regression, simple neural networks. | Uses Genetic Programming to evolve pipelines with crossover/mutation for objective metric improvement. |

Abbreviations: CV, Cross-Validation; DNN, Deep Neural Network; GLM, Generalized Linear Model; KNN, K-Nearest Neighbors, MLP, Multilayer Perceptron; RFE, Recursive Feature Elimination; SGD; Stochastic Gradient Descent; SHAP, Shapley Additive exPlanations; SVM, Support Vector Machine, XRT, eXtra Random Trees.

# References

[1]     J. J. M. van Griethuysen *et al.*, "Computational Radiomics System to Decode the Radiographic Phenotype," *Cancer Research*, vol. 77, no. 21, pp. e104–e107, Nov. 2017, doi: 10.1158/0008-5472.CAN-17-0339.